\documentclass[11pt]{article}
\usepackage{multirow}
\usepackage{rotating}
\usepackage{eacl2017}
\usepackage{times}
\usepackage{latexsym}
\usepackage{url}
\usepackage{arydshln}
\usepackage{graphicx}
\usepackage{amsmath}
\usepackage{amsfonts, amssymb}
\usepackage{xcolor}

\def\newparagraph#1{\textbf{#1.}}

\newcounter{notecounter}

\newcommand{\enoteson}{\long\gdef\enote##1##2{{
\stepcounter{notecounter}
{\large\bf
\hspace{1cm}\arabic{notecounter} $<<<$ ##1: ##2
$>>>$\hspace{1cm}}}}}
\enoteson

\eaclfinalcopy 

\long\def\eat#1{\ignorespaces}

\long\def\onlyincludeinlongversion#1{#1}
\long\def\onlyincludeinshortversion#1{\ignorespaces}

\def\mydelimit{$\parallel$}
\def\dnrm#1{\mbox{$_{\hbox{\scriptsize #1}}$}}

\def\figref#1{Figure~\ref{fig:#1}}
\def\figlabel#1{\label{fig:#1}\label{p:#1}}

\def\tabref#1{Table~\ref{tab:#1}}
\def\tablabel#1{\label{tab:#1}\label{p:#1}}

\def\secref#1{\S\ref{sec:#1}}
\def\seclabel#1{\label{sec:#1}\label{p:#1}}
\def\eqref#1{Eq.~\ref{eqn:#1}}

\title{Nonsymbolic Text Representation\onlyincludeinlongversion{$^{\textbf{0}}$}}
\author{Hinrich Sch\"{u}tze, Heike Adel, Ehsaneddin Asgari\\ CIS, LMU Munich, Germany\\{\tt
    inquiries@cislmu.org}}
\date{}
\begin{document}

\maketitle

\begin{abstract}
We introduce the first generic text representation model that is
completely nonsymbolic, i.e., it does not require the
availability of a segmentation or tokenization method that
attempts to identify words or other symbolic units in text.
This applies to training the 
representations as well as to using them in an application.
We demonstrate better performance  than prior work
on entity typing and text denoising.
\end{abstract}

\section{Introduction}
\seclabel{intro}
Character-level models can be grouped into three classes.
(i) \textbf{End-to-end models}
learn a separate model on the raw character (or byte) input
for each task; these models estimate task-specific
parameters, but no representation of text that would be
usable across tasks is computed. Throughout this paper, 
we refer to $r(x)$ as the ``representation'' of $x$ only if
$r(x)$ is a generic rendering of $x$ that can be used in a
general way, e.g., across tasks and domains. 
The activation pattern of a hidden layer
for a given input
sentence in a multilayer perceptron (MLP) is not a representation according to this
definition if it  is not used outside of the MLP.
(ii)
\textbf{Character-level models of words} derive a
representation of a word $w$
from the character string of $w$, but they are symbolic in that
they need text segmented into tokens as input. (iii) 
Bag-of-character-ngram models, 
\textbf{bag-of-ngram models} for short, use character ngrams to
encode sequence-of-character information, but
sequence-of-ngram information is lost in the representations
they produce.\onlyincludeinlongversion{\footnotetext[0]{A short version of this paper
  appears as \cite{schuetze_2017_EACL}.}}

Our premise is that text representations are
needed in NLP.  A large body of work on  word
embeddings  
demonstrates that a generic text representation, trained in
an unsupervised fashion on large corpora, is useful. Thus, we take the view that \textbf{group (i) models},
end-to-end learning without any representation learning, is
not a good general approach for NLP.

We distinguish \textbf{training}
and  \textbf{utilization}
of the text representation model. We use ``training''
to refer to the method by which the model is learned and
``utilization'' to refer to the application of the model to a piece
of text to compute a representation of the text. 
In many text representation models, utilization is
trivial. For example, for word embedding models, utilization
amounts to a simple lookup of a word to get its precomputed
embedding. However, for the models we consider, utilization
is not trivial and we will discuss different approaches.

Both training and utilization can be  either \textbf{symbolic}
or \textbf{nonsymbolic}. We define a symbolic approach as one
that is based on tokenization, i.e., a segmentation of the
text into tokens. Symbol identifiers (i.e., tokens) can 
have internal structure -- a tokenizer may recognize
tokens like ``to and fro'' and ``London-based'' that contain
delimiters --  and may be morphologically analyzed
downstream.\footnote{\onlyincludeinlongversion{Two comments on how
    we use the term ``symbolic''. (i) Some define  symbols  as identifiers
  that have no internal structure, but uses of the term that
  allow for the possibility of internal structure are also
  frequent. For example, the Wikipedia entry for
  ``symbol'' states: ``Numerals are symbols for numbers.''
  Most numerals have internal structure. (ii) }The
  position-embedding representation of a text introduced
  below is a sequence
  of position embeddings. 
An embedding that represents a single character must be
viewed as
symbolic since a character is a symbol -- just
like a representation of text as a sequence of word
embeddings is symbolic since each word corresponds to a symbol. But
position embeddings do not
  represent single characters. See \secref{utilize}.}

We define a nonsymbolic approach as one that is
tokenization-free, i.e., no assumption is made that there
are segmentation boundaries and that each segment (e.g., a
word) should be represented 
(e.g., by a word embedding) in a way that is independent of
the representations (e.g., word embeddings) of neighboring segments.
Methods for training
text representation models 
that require tokenized text
include
word embedding models 
like word2vec \cite{mikolov13word2vec}
and 
most \textbf{group (ii) methods}, i.e.,
character-level models like
fastText skipgram \cite{bojanowski17enriching}.

\textbf{Bag-of-ngram models}, group (iii) models,
are text representation utilization
models that typically compute the
representation of a text as the sum of the embeddings of all character ngrams
occurring in it, 
e.g., WordSpace \cite{schutze92wordspace} and CHARAGRAM
\cite{wieting16charagram}. WordSpace and CHARAGRAM are
examples of
mixed training-utilization models: training is performed on
tokenized text (words and
phrases), \onlyincludeinlongversion{\footnote{The WordSpace model could be trained on text that is
    not tokenized, but the redundancy of overlapping
    character ngrams could be a problem: see discussion below.}}
utilization is nonsymbolic. 

We make two contributions in this paper.
(i)
We propose
the first generic 
method for training text representation models without the
need for tokenization and address 
the challenging sparseness issues that make this  difficult.
(ii)
We
propose the first nonsymbolic utilization method 
that fully represents sequence information -- in contrast to
utilization methods like
bag-of-ngrams that discard sequence information 
that is not
directly encoded in the character ngrams themselves.

\section{Motivation}
\newcite{chung16characternmt} give two  motivations
for their work
on character-level models. First, 
\textbf{tokenization} (or, equivalently,
segmentation) \textbf{algorithms make many mistakes} and are brittle:
``we do not have a perfect word segmentation algorithm for
any one language''. Tokenization errors then propagate
throughout the NLP pipeline.

Second, there is currently 
\textbf{no general solution for morphology} in statistical
NLP. For many languages, high-coverage and high-quality 
morphological resources are not available. Even for well
resourced languages, problems like ambiguity make
morphological processing difficult; e.g., 
``rung'' is either the singular of a noun meaning
``part of a ladder'' or the past participle of ``to
ring''. In many languages, e.g., in German, syncretism,
a particular type of systematic morphological ambiguity, is
pervasive. Thus, there is no simple 
morphological processing method that would produce a
representation in which all inflected forms of ``to ring''
are marked as having a common lemma; and no such method in which an unseen
form like
``aromatizing'' is reliably analyzed as a form of
``aromatize'' whereas an unseen form like ``antitrafficking'' is
reliably analyzed as the compound ``anti+trafficking''.

Of course, it is an open question whether nonsymbolic
methods can perform better than morphological analysis, but
the foregoing discussion  motivates us to investigate them.

\newcite{chung16characternmt} focus on problems with 
the tokens produced by segmentation algorithms.
Equally important is the problem that \textbf{tokenization fails to
capture structure across multiple tokens}. The job of dealing
with cross-token structure is often given to downstream components
of the pipeline, e.g., components that recognize 
multiwords and named entitites in English or in fact any
word in a language like Chinese that uses no overt
delimiters. However, there is no linguistic
or computational reason in principle why we should treat the recognition
of a unit like ``electromechanical'' (containing no space) as fundamentally different
from the recognition of a unit like ``electrical engineering''
(containing a space). Character-level models offer the
potential of uniform treatment of such linguistic units.

\section{Text representation model: Training}

\subsection{Methodology}
\seclabel{methodology}
Many text representation learning algorithms can be
understood as estimating the parameters of the model from a
unit-context matrix $C$ where each row corresponds to a unit
$u_i$, each column to a context $c_j$ and each cell $C_{ij}$
measures the degree of association between
$u_i$ and $c_j$.
For example, the skipgram model is closely
related to an SVD factorization of a pointwise
mutual information  matrix
\cite{levy14neural2}\onlyincludeinlongversion{; in this case, both units and contexts
are words}.
Many text representation learning algorithms are
formalized as matrix factorization (e.g.,
\cite{deerwester90indexing2,hofmann99plsi2,stratos15modelbased2}),
but there may be no big
difference between implicit 
(e.g., \cite{pennington14glove2})
and explicit factorization methods;
see also \cite{mohamed11factorization2,rastogi2015multiview2}.

Our goal in this paper is not to develop new 
matrix factorization
methods. Instead, we will focus on defining the
unit-context matrix in such a way that no symbolic
assumption has to be made. This unit-context matrix can then
be processed by any existing or still to be invented
algorithm.

\textbf{Definition of units and contexts.}
How to define units and contexts without relying on
segmentation boundaries?  In initial experiments, we simply
generated all character ngrams of length up to $k\dnrm{max}$ (where
$k\dnrm{max}$ is a parameter),
including character ngrams that cross token
boundaries; i.e., no segmentation is needed. We then used a skipgram-type
objective for learning
embeddings that attempts to
predict, from
ngram $g_1$,  an ngram $g_2$ in $g_1$'s
context. Results were poor
because many training instances
consist of pairs $(g_1,g_2)$ in which $g_1$ and $g_2$
overlap, e.g., one is a subsequence of the other. So the
objective encourages trivial predictions of ngrams that have
high string similarity with the input and nothing
interesting is learned.

In this paper, we propose an alternative way of
defining units and
contexts that supports well-performing nonsymbolic text representation
learning: \textbf{multiple random segmentation}. A pointer
moves through the training corpus.  The current position $i$
of the pointer defines the left boundary of the next
segment.  The length $l$ of the next move is uniformly
sampled from $[k\dnrm{min},k\dnrm{max}]$ where $k\dnrm{min}$
and $k\dnrm{max}$ are the minimum and maximum segment lengths.
The right boundary of the segment is then
$i+l$. Thus, the segment just generated is $c_{i,i+l}$, the
subsequence of the corpus between (and including) positions
$i$ and $i+l$. The pointer is positioned at $i+l+1$, the
next segment is sampled and so on. An example of a random
segmentation from our experiments is 
``@he@had@b egu n@to@show @his@cap acity@f'' where space was
replaced with ``@'' and the next segment starts with ``or@''.

The corpus is segmented this way $m$ times (where $m$ is a
parameter) and the $m$ random segmentations are
concatenated. The unit-context matrix is derived from this
concatenated corpus.

Multiple random segmentation has two advantages. First,
there is no redundancy since, in any given random
segmentation, two ngrams do not overlap and are not
subsequences of each other. Second, a single random
segmentation would only cover a small part of the space of
possible ngrams. For example, a random segmentation of ``a
rose is a rose is a rose'' might be ``[a ros][e is a ros][e
  is][a rose]''. This segmentation does not contain the
segment ``rose'' and this part of the corpus can then not be
exploited to learn a good embedding for the fourgram
``rose''. However, with multiple random segmentation,
it
is likely that this part of the corpus does give rise to the
segment ``rose'' in one of the segmentations and can
contribute information to learning a good embedding for
``rose''.

We took the idea of random segmentation from work on
biological sequences
\cite{asgari15biological,asgari16fifty}. 
Such sequences have no delimiters, so they are a
good model if one believes that delimiter-based segmentation
is problematic for text.

\subsection{Ngram equivalence
classes/Permutation}
\seclabel{permutation}

\textbf{Form-meaning homomorphism premise.} 
Nonsymbolic representation learning
does not preprocess the training corpus by means of
tokenization and considers many ngrams that would be ignored
in tokenized approaches because they span token
boundaries. As a result, the number of ngrams that occur in a corpus is an
order of magnitude larger for tokenization-free approaches
than for tokenization-based approaches. See supplementary
for details.

We will see below that this sparseness impacts performance
of nonsymbolic text representation negatively. We address
sparseness by defining ngram equivalence classes. All ngrams
in an equivalence class receive the same embedding.

The relationship between form and meaning is mostly
arbitrary, but there are substructures of the ngram space
and the embedding space that are systematically related
by homomorphism. In this paper, \textbf{we will assume the following homomorphism:}
\[
g_1 \sim_{\tau} g_2 \Leftrightarrow \vec{v}(g_1) \sim_{=} \vec{v}(g_2)
\]
where
$g_1 \sim_{\tau} g_2$ iff $\tau(g_1)\!\!=\!\!\tau(g_2)$ for  string transduction $\tau$
and $\vec{v}(g_1) \sim_{=} \vec{v}(g_2)$ iff $|\vec{v}(g_1)-\vec{v}(g_2)|_2<\epsilon$.

As a simple example 
consider a transduction $\tau$ that deletes spaces at the beginning of ngrams,
e.g., 
$\tau(\mbox{@Mercedes}) = \tau(\mbox{Mercedes})$. This is an
example of a
meaning-preserving $\tau$ since for, say, English, 
$\tau$ will not change meaning.
We will propose a procedure for learning $\tau$
below.

We define $\sim_{=}$ as ``closeness'' -- not as identity --
because of estimation noise when embeddings are learned. We
assume that 
there are no true synonyms and therefore  the
direction
$g_1 \sim_{\tau} g_2
\Leftarrow 
\vec{v}(g_1) \sim_{=} \vec{v}(g_2)$
also holds.
For example,
``car'' and ``automobile'' are considered synonyms, but we
assume that their embeddings are different because
only ``car'' has the literary sense ``chariot''.
If
they were identical, then the homomorphism would not hold
since ``car'' and ``automobile'' cannot be converted
into each other by any plausible meaning-preserving $\tau$.

\textbf{Learning procedure.} 
To learn $\tau$, we define three
templates that transform one ngram into another: (i) replace
character $a_1$ with character $a_2$, 
(ii) delete character
$a_1$ if its immediate predecessor is character $a_2$, 
(iii) delete character
$a_1$ if its immediate successor is character $a_2$.
The learning procedure takes a set of ngrams and their embeddings as
input. It then exhaustively searches for all pairs of ngrams,
for all pairs of characters $a_1$/$a_2$, for each of the
three templates. \onlyincludeinlongversion{(This
takes about 10 hours on a multicore server.)}
When two matching embeddings exist, we compute their cosine.
For example, for the operation ``delete space before M'',
an ngram pair from our embeddings that matches
is ``@Mercedes''
/ ``Mercedes'' and we compute its cosine.
As the characteristic statistic of an operation
we take the average of all cosines; e.g., 
for ``delete space before M'' the average cosine is
.7435. We then rank operations according to average cosine
and take the first $N_o$ as the definition of $\tau$ where
$N_o$ is a parameter. For characters
that are replaced by each other (e.g., 1, 2, 3
in \tabref{rules}),
we compute the equivalence class and then replace the learned
operations with ones that replace a character by the
canonical member of its equivalence class (e.g., 
2 $\rightarrow$ 1, 3 $\rightarrow$ 1).

\textbf{Permutation premise.}  Tokenization algorithms can
be thought of as assigning a particular function or
semantics to each character and making tokenization
decisions accordingly; e.g., they may disallow that a semicolon, the
character ``;'', occurs inside a token. If we
want to learn representations from the data without imposing
such hard constraints, then characters should not have any
particular function or semantics. A consequence of this
desideratum is that if any two characters are exchanged for
each other, this should not affect the representations that
are learned. For example, if we interchange space and ``A''
throughout a corpus, then this should have no effect on
learning: what was the representation of ``NATO'' before,
should now be the representation of ``N TO''.
We can also think of this type of permutation as a
sanity check: it ensures we do not inadvertantly make use of
text preprocessing heuristics that are pervasive in
NLP.\footnote{An example of such an inadvertant use of
  text preprocessing heuristics is that fastText seems to default
  to lowercase ngrams if embeddings of uppercase ngrams are not
  available: when fastText is trained on lowercased text and
  then applied to uppercased text, it still produces embeddings.}

Let $A$ be the alphabet of a language, i.e., its set of
characters, $\pi$ a permutation on $A$, $C$ a corpus and
$\pi(C)$ the corpus permuted by $\pi$.
For example, if $\pi(a) = e$, then all
 ``a'' in $C$ are replaced with ``e'' in 
$\pi(C)$. \textbf{The learning procedure should learn
identical equivalence classes on 
$C$ and $\pi(C)$.} So, if $ g_1 \sim_{\tau} g_2$ 
after running
the
learning procedure on $C$, then 
 $ \pi(g_1) \sim_{\tau} \pi(g_2)$ after running the learning
procedure on 
$\pi(C)$. 

This premise is motivated by our desire to come up with a
general method that does not rely on specific properties of
a language or genre; e.g., the premise rules out
exploiting the fact through feature engineering that in many languages and genres, ``c''
and ``C'' are related. Such a relationship has to be learned
from the data.

\subsection{Experiments}
We run experiments on $C$, a 3 gigabyte
English Wikipedia corpus, and train word2vec skipgram (W2V, \cite{mikolov13word2vec})
and fastText skipgram (FTX, \cite{bojanowski17enriching}) models on $C$ and its derivatives.
We randomly generate a permutation $\pi$ on the alphabet and
learn a transduction $\tau$ (details below).
In \tabref{figment} (left), the columns ``method'', $\pi$ and
$\tau$ indicate the method used (W2V or FTX) and 
whether experiments in a row were run on $C$, $\pi(C)$ or
$\tau(\pi(C))$. The values of ``whitespace'' are: (i)
ORIGINAL (whitespace as in the original), (ii) SUBSTITUTE (what
$\pi$ outputs as whitespace is used as whitespace, 
i.e.,
$\pi^{-1}(\mbox{`` ''})$ becomes the new whitespace)  and
  (iii) RANDOM (random segmentation with parameters $m=50$, $k\dnrm{min}=3$, 
$k\dnrm{max}=9$). Before random
  segmentation, whitespace is replaced with ``@'' -- this
  character occurs rarely in $C$, so that the effect of
  conflating two characters (original ``@'' and whitespace) can be neglected.
The random segmenter then indicates boundaries by whitespace
-- unambiguously since it is applied to text that contains no whitespace.

We learn $\tau$ on the embeddings learned by 
W2V on the random segmentation version of $\pi(C)$
(C-RANDOM in the table) as described in \secref{permutation}
for $N_o=200$.
Since the number of equivalence classes is
much smaller than the number of ngrams, $\tau$
reduces the number of distinct
character ngrams from 758M in the random segmentation
version of $\pi(C)$ (C/D-RANDOM) to 96M in the random segmentation
version of $\tau(\pi(C))$ (E/F-RANDOM).

\begin{table}
\begin{tabular}{l@{\hspace{0.3cm}}l@{\hspace{0cm}}l@{\hspace{0cm}}l|l@{\hspace{0.3cm}}l@{\hspace{0cm}}l@{\hspace{0cm}}l|l@{\hspace{0.3cm}}l@{\hspace{0cm}}l@{\hspace{0cm}}l}
\multirow{6}{*}{\begin{turn}{90}{\small substitution}\end{turn}} & 2 & $\rightarrow$ & 1&
\multirow{5}{*}{\begin{turn}{90}{\small predeletion}\end{turn}}& /r & $\rightarrow$ & r&
\multirow{5}{*}{\begin{turn}{90}{\small postdeletion}\end{turn}}& $\ddagger$@ & $\rightarrow$ & $\ddagger$\\
 & 3 & $\rightarrow$ & 1&
& @$\ddagger$ & $\rightarrow$ & $\ddagger$&
& e@ & $\rightarrow$ & e\\
 & : & $\rightarrow$ & .&
& @$\ddagger$ & $\rightarrow$ & $\ddagger$&
& l@ & $\rightarrow$ & l\\
 & ; & $\rightarrow$ & .&
& @H & $\rightarrow$ & H&
& m@ & $\rightarrow$ & m\\
 & E & $\rightarrow$ & e&
& @I & $\rightarrow$ & I&
& ml & $\rightarrow$ & m\\
 & C & $\rightarrow$ & c
\end{tabular}

\caption{String operations that on average do not change
  meaning. ``@'' stands for space.
$\ddagger$ is the left or right boundary of the ngram.
  \tablabel{rules}}
\end{table}

\tabref{rules} 
shows a selection of the $N_o$ operations.
Throughout the paper, if we give examples from $\pi(C)$ or
$\tau(\pi(C))$ as we do here, 
we  convert characters back to the original for better readability.
The two
uppercase/lowercase conversions shown in the table 
(E$\rightarrow$e, 
C$\rightarrow$c)
were the
only ones that were learned (we had hoped for more). The
postdeletion rule ml$\rightarrow$m usefully rewrites  ``html''
as ``htm'', but is likely to do more harm than good.
We inspected all
200 rules and, with a few exceptions like
ml$\rightarrow$m, they looked good to us.

\textbf{Evaluation.}
We evaluate the three models on an
entity typing task, similar to 
\cite{yaghoobzadeh15typing2}, but
 based on an entity dataset released by
\newcite{xie16entitydesc2} in which each entity has been assigned
one or more types from a set of 50 types. 
For example,
the entity ``Harrison Ford''
has the types
``actor'', ``celebrity'' and ``award winner'' among
others.
We extract mentions from FACC 
(\url{http://lemurproject.org/clueweb12/FACC1})
if an entity has
a mention there or we use the Freebase name as the mention
otherwise. 
This gives us a data set of 
54,334, 6085 and
6747 mentions in train, dev and test, respectively.
Each mention is annotated with the types that its entity has
been assigned by \newcite{xie16entitydesc2}. 
The evaluation has a strong cross-domain aspect because of
differences between FACC and Wikipedia, the training corpus for our
representations.
For example, of the 525 mentions in dev that have a length
of at least 5 and do not contain lowercase characters, more
than half have 0 or 1 occurrences in the Wikipedia corpus, including
many like ``JOHNNY CARSON'' that are frequent in other case
variants.

\begin{table*}
\begin{tabular}{l@{\hspace{0.05cm}}l}
\footnotesize{
\begin{tabular}{l@{\hspace{0.1cm}}l|ll|l||l@{\hspace{0.05cm}}l@{\hspace{0.05cm}}l|l@{\hspace{0.05cm}}l@{\hspace{0.05cm}}l|l@{\hspace{0.05cm}}l@{\hspace{0.05cm}}l}
& &     & &\tiny{whitespace} & \multicolumn{3}{c|}{\small{ORIGINAL}}     &\multicolumn{3}{c|}{\small{SUBSTITUTE}}     &
  \multicolumn{3}{c}{\small{RANDOM}}\\\hline
&&&&\tiny{measure}&\multicolumn{1}{c}{$P$}&\multicolumn{1}{c}{$R$}&\multicolumn{1}{c|}{$F_1$}&\multicolumn{1}{c}{$P$}&\multicolumn{1}{c}{$R$}&\multicolumn{1}{c|}{$F_1$}&\multicolumn{1}{c}{$P$}&\multicolumn{1}{c}{$R$}&\multicolumn{1}{c}{$F_1$}\\\hline
&\tiny{method}&$\pi$&$\tau$ &     &&&&&&\\\hline\hline
A&W2V&$-$ &$-$    &&     .538 &.566 &.552 &&&&.525& .596 &.558\\ 
B&FTX&$-$   &$-$   &     & .530 &.628 &.575 &&&&.528 & .608 & .565 \\\hline
C&W2V&$+$  &$-$   &&.535 &.560 &.547& .191 & .296 & .233 &.514 &.605 & .556 \\
D&FTX&$+$ &$-$     && .530 & .623 & .573 &.335 & .510 &.405&  .531 &.608 &.567\\\hline
E&W2V&$+$  &$+$   &     &&&&&&&.503 &.603 &.548  \\
F&FTX&$+$   &$+$  &     &&&&&&&.551 & .618 & .582 
\end{tabular}
}
&
\scriptsize{
\begin{tabular}{r@{\hspace{0.3cm}}l@{\hspace{0.3cm}}l@{\hspace{0.3cm}}l}
&query & neighbor & $r$\\\hline
 1&\texttt{Abdulaziz} & \texttt{Abdul Azi} & 2\\
 2&\texttt{codenamed} & \texttt{code name} & 1\\
 3&\texttt{Quarterfi} & \texttt{uarter-Fi} & 1\\
 4&\texttt{worldreco} & \texttt{orld-reco} & 1\\\hline
 5&\texttt{antibodie} & \texttt{stem cell} &1\\
 6&\texttt{eflectors} & \texttt{ear wheel} & 1\\
 7&\texttt{ommandeer} & \texttt{rash land} & 1\\
 8&\texttt{reenplays} & \texttt{ripts for} & 1\\
 9&\texttt{roughfare} & \texttt{ugh downt} & 1\\\hline
10&\texttt{ilitating} & \texttt{e-to-face} & 1
\end{tabular}}

\end{tabular}
\caption{Left: Evaluation results for named entity typing.
Right: Neighbors of character ngrams.  Rank $r=1$/$r=2$:
  nearest / second-nearest neighbor.
  \tablabel{figment}}
\end{table*}

Since our goal in this experiment is to evaluate
tokenization-free learning, not tokenization-free
utilization, we use a simple utilization baseline, the
bag-of-ngram model (see \secref{intro}). A mention is
represented as the sum of all character ngrams that
embeddings were learned for.  Linear SVMs \cite{libsvm2} are
then trained, one for each of the 50 types, on train and
applied to dev and test.  Our evaluation measure is micro
$F_1$ on all typing decisions; e.g., one typing decision is:
``Harrison Ford'' is a mention of type ``actor''.  We tune
thresholds on dev to optimize $F_1$ and then use these
thresholds on test.

\subsection{Results}
Results are presented in \tabref{figment} (left). Overall
performance of FTX is higher than W2V in all cases. For
ORIGINAL, FTX's recall is a lot higher than W2V's whereas
precision decreases slightly. This indicates that FTX is
stronger in both learning and application: in learning it
can generalize better from sparse training data and in
application it can produce  representations for OOVs and
better representations for rare words. For English,
prefixes, suffixes and stems are of particular importance,
but 
there
often is not a neat correspondence between these traditional
linguistic concepts and internal FTX representations; e.g.,
\newcite{bojanowski17enriching} show that ``asphal'',
``sphalt'' and ``phalt'' are informative character ngrams of
``asphaltic''.

Running
W2V on random
segmentations can be viewed as an alternative to the
learning mechanism of FTX, which is based on character ngram
cooccurrence; so it is not surprising that 
for RANDOM, FTX has only a small advantage over W2V. 

For C/D-SUBSTITUTE, we see a dramatic loss in performance if
tokenization heuristics are not used. This is not
surprising, but shows how powerful 
tokenization can be.

C/D-ORIGINAL  is like C/D-SUBSTITUTE except
that we artificially restored the space -- so the
permutation $\pi$ is applied to all characters except for
space. By comparing
C/D-ORIGINAL  and C/D-SUBSTITUTE, we see that the space is
the most important text preprocessing feature employed by W2V and FTX. If
space is restored, there is only a small loss of performance
compared to A/B-ORIGINAL. So text preprocessing heuristics
other than whitespace tokenization in a narrow definition of
the term (e.g., downcasing) do not seem to play a big role,
at least not for our entity typing task.

For tokenization-free embedding learning on random
segmentation, there is almost no difference between original
data (A/B-RANDOM) and permuted data (C/D-RANDOM). This
confirms that our proposed learning method is insensitive to
permutations and makes no use of text preprocessing heuristics.

We achieve an additional improvement by applying the
transduction $\tau$. In fact, FTX performance for F-RANDOM
($F_1$ of .582) is better than tokenization-based W2V and
FTX performance. Thus, our proposed method seems to be an
effective tokenization-free alternative to
tokenization-based embedding learning.

\subsection{Analysis of ngram embeddings}
\tabref{figment} (right) shows nearest neighbors of ten character
ngrams, for the A-RANDOM space. 
Queries were chosen to contain only alphanumeric
characters. To highlight the difference to symbol-based
representation models, we restricted the search to
9-grams that contained a delimiter 
at positions 3, 4, 5, 6 or 7.

Lines 1--4 show that ``delimiter variation'',
i.e., cases where a word has two forms, one with a
delimiter, one without a delimiter,
is handled well:
``Abdulaziz'' / ``Abdul Azi'',
``codenamed'' / ``code name'',
``Quarterfinal'' / ``Quarter-Final'', ``worldrecord'' / ``world-record''.

Lines 5--9 are cases of ambiguous or polysemous
words that are disambiguated through ``character context''. ``stem'', ``cell'', ``rear'', ``wheel'', ``crash'',
``land'', ``scripts'', ``through'', ``downtown'' all have
several meanings. In contrast, the meanings of ``stem
cell'', ``rear wheel'', ``crash land'', ``(write) scripts
for'' and ``through downtown'' are 
less ambiguous. 
A multiword recognizer may find the phrases
``stem cell'' and ``crash land'' automatically. But
the examples of ``scripts for'' and ``through downtown''
show that what is accomplished here is not multiword
detection, but a more general use of character context for
disambiguation.

Line 10 shows that a 9-gram of ``face-to-face'' is the
closest neighbor to a 9-gram of ``facilitating''. This
demonstrates that form and meaning sometimes interact in
surprising ways. Facilitating a meeting is most commonly
done face-to-face. It is not inconceivable that form -- the
shared trigram ``fac'' or the shared fourgram ``faci'' in
``facilitate'' / ``facing'' -- is influencing meaning here
in a way that also occurs historically in cases like ``ear''
`organ of hearing' / ``ear'' `head of cereal plant',
originally unrelated words that many 
English speakers today intuit as one word.

\section{Utilization: Tokenization-free representation of text}
\seclabel{utilize}

\subsection{Methodology}
The main text representation model that is based on ngram
embeddings similar to ours is the \textbf{bag-of-ngram model}.  A
sequence of characters is represented by a single vector
that is computed as the sum of the embeddings of all ngrams
that occur in the sequence. 
In fact, this is what we did in the entity typing experiment.
In most work on bag-of-ngram
models, the sequences considered are words or phrases\onlyincludeinshortversion{ (see
\cite{schuetze16nonsymboliclong2} for citations)}. 
In a few cases, the model
is applied to longer sequences, including sentences and
documents; e.g., \cite{schutze92wordspace}, \cite{wieting16charagram}.

The basic assumption of the bag-of-ngram model is that
sequence information is encoded in the character ngrams and
therefore a ``bag-of'' approach (which usually throws away
all sequence information) is sufficient.
The assumption is
not implausible: for most bags of character sequences,
there is only a single way of stitching them together to one
coherent sequence, so in that case information is not necessarily
lost (although this is likely when embeddings are added).
But the assumption has not been tested experimentally.

\begin{table*}
\footnotesize{
\begin{tabular}{rl|llllll}
 \multicolumn{1}{c}{POS} &          & $r=1$& $r=2$& $r=3$& $r=4$& $r=5$\\\hline
2&\texttt{e}&\texttt{wealthies} & \texttt{accolades} & \texttt{bestselle} & \texttt{bestselli} & \texttt{Billboard}\\
3&\texttt{s}&\texttt{estseller} & \texttt{wealthies} & \texttt{bestselli} & \texttt{accolades} & \texttt{bestselle}\\
15&\texttt{o}&\texttt{fortnight} & \texttt{afternoon} & \texttt{overnight} & \texttt{allowance} & \texttt{Saturdays}\\
16&\texttt{n}&\texttt{fortnight} & \texttt{afternoon} & \texttt{Saturdays} & \texttt{Wednesday} & \texttt{magazines}\\
23&\texttt{o}&\texttt{superhero} & \texttt{ntagraphi} & \texttt{adventure} & \texttt{Astonishi} & \texttt{bestselli}\\
24&\texttt{m}&\texttt{superhero} & \texttt{ntagraphi} & \texttt{anthology} & \texttt{Daredevil} & \texttt{Astonishi}\\
29&\texttt{o}&\texttt{anthology} & \texttt{paperback} & \texttt{superhero} & \texttt{Lovecraft} & \texttt{tagraphic}\\
30&\texttt{o}&\texttt{anthology} & \texttt{paperback} & \texttt{tagraphic} & \texttt{Lovecraft} & \texttt{agraphics}\\
34&\texttt{u}&\texttt{antagraph} & \texttt{agraphics} & \texttt{paperback} & \texttt{hardcover} & \texttt{ersweekly}\\
35&\texttt{b}&\texttt{ublishing} & \texttt{ublishers} & \texttt{ublicatio} & \texttt{antagraph} & \texttt{aperbacks}
\end{tabular}
}
\caption{\tablabel{visu}
Nearest ngram embeddings (rank $r \in [1,5]$) of the position embeddings 
for ``POS'', the positions 2/3 (b\underline{es}t), 
15/16 (m\underline{on}thly), 23/24 (c\underline{om}ic), 29/30
(b\underline{oo}k) and 34/35
(p\underline{ub}lications) 
in the Wikipedia excerpt ``best-selling monthly comic book publications sold in North America''
}
\end{table*}

Here, we propose \textbf{position embeddings},
character-ngram-based embeddings that more fully preserve
sequence information.\footnote{Position embeddings were
  independently proposed by
\newcite{kalchbrenner16lineartime}, see Section~3.6 of their
paper.} The simple idea is to represent each
position as the sum of all ngrams that contain that
position. When we set $k\dnrm{min}=3$, $k\dnrm{max}=9$,
this means that the position is the
sum of $(\sum_{3 \leq k \leq 9} k)$ ngram embeddings
(if all of these ngrams have embeddings, which generally will 
be true for some, but
not
for most positions). A sequence of $n$ characters is then
represented as a sequence of $n$ such position embeddings.

\subsection{Experiments}
We again use the embeddings corresponding to A-RANDOM in \tabref{figment}.
We randomly selected  2,000,000 contexts of size 40 characters
from Wikipedia. We then created a noise context
for each of the 2,000,000 contexts by replacing one
character at position i ($15\leq i \leq
25$, uniformly sampled) with space (probability $p= .5$) or a random
character otherwise. Finally, we selected 1000 noise contexts
randomly and computed their nearest neighbors among the
4,000,000 contexts (excluding the noise query). We did this in two different conditions:
for a bag-of-ngram representation of the context (sum of all
character ngrams) and for the concatenation
of 11 position embeddings, those between 15 and
25. Our evaluation measure is mean reciprocal rank of the
clean context corresponding to the noise context. This
simulates a text denoising experiment: if the clean
context has rank 1, then the noisy context can be corrected.

\begin{table}[b]
\begin{tabular}{l|ll}
& bag-of-ngram & position embeddings\\\hline
MRR & .64 & .76
\end{tabular}
\caption{Mean reciprocal rank of text denoising experiment
  for bag-of-ngram text representation and
  position embedding text representation
  \tablabel{mrr}}
\end{table}

\tabref{mrr} shows that sequence-preserving position embeddings
perform better than bag-of-ngram
representations.

\begin{table*}
\small{
\begin{tabular}{l@{\hspace{0.15cm}}rllll@{\hspace{0cm}}l@{\hspace{0cm}}l@{\hspace{0cm}}l@{\hspace{0cm}}lllll}
& &rep.\ space& similarity & $r$ & \multicolumn{1}{c}{left context} &  \multicolumn{1}{c}{center} & \multicolumn{1}{c}{right context}\\\hline\hline
1&correct    &&& &\mydelimit\texttt{s and Seattle S}&\mydelimit\texttt{eahawks th}&\mydelimit\texttt{at led to publi}&\mydelimit\\\hline
2&noise (query)&     && &\mydelimit\texttt{s and Seattle S}&\mydelimit\texttt{eahawks t }&\mydelimit\texttt{at led to publi}&\mydelimit\\\hline
3&&position-emb
&.761   &1&\mydelimit\texttt{s and Seattle S}&\mydelimit\texttt{eahawks th}&\mydelimit\texttt{at led to publi}&\mydelimit\\\hline
4&&bag-of-ngram &.904 &1&\mydelimit\texttt{arted 15 games}&\mydelimit\texttt{fsr the Se}&\mydelimit\texttt{ahawks, leading}&\mydelimit\\
5&&bag-of-ngram &.864 &6&\mydelimit\texttt{s and Seattle S}&\mydelimit\texttt{eahawks th}&\mydelimit\texttt{at led to publi}&\mydelimit\\\hline
\end{tabular}}
\caption{Illustration of the result in \tabref{mrr}. 
``rep.\ space'' = ``representation space''.
We want
  to correct the error in the corrupted ``noise'' context
  (line 2) and produce ``correct'' (line 1). The nearest
  neighbor to line 2 in position-embedding space is the
  correct context (line 3, $r=1$).
 The nearest
  neighbor to line 2 in bag-of-ngram space is incorrect
  (line 4, $r=1$)
because the precise position of ``Seahawks'' in
  the query is not encoded.
The correct context in bag-of-ngram space is instead at rank
$r=6$ (line 5).
``similarity'' is average cosine (over eleven position
embeddings) for position embeddings.
  \tablabel{knn}}
\end{table*}

\tabref{knn} shows an example of a context in which position
embeddings did better than bag-of-ngrams, demonstrating
that  sequence information is lost
by bag-of-ngram representations, in this case the exact
position of ``Seahawks''.

\tabref{visu}
gives further intuition about the type of information
position embeddings contain, 
showing the ngram embeddings
closest to selected position embeddings; e.g., ``estseller'' (the first 9-gram
on the line numbered 3 in the table)
is closest to the
embedding of position 3 (corresponding to the first ``s'' of ``best-selling''). The kNN search space is restricted to
alphanumeric ngrams.

\def\myph{\phantom{.}}

\begin{table}

{\scriptsize
\begin{tabular}{l|rrr}
&exchange@f&ic@exchang&ing@exchan\\
& (in exchange for) & (many contexts) & (many contexts)\\\hline 
exchange@f&     1.000&     0.008&    -0.056\\
ic@exchang&     0.008&     1.000&     0.108\\
ing@exchan&    -0.056&     0.108&     1.000\\\hline\hline
&xchange@ra&ival@rates&rime@rates\\
& (exchange rates) & (survival rates) & (crime rates)\\\hline
xchange@ra&     1.000&     0.036&     0.050\\
ival@rates&     0.036&     1.000&     0.331\\
rime@rates&     0.050&     0.331&     1.000
\end{tabular}
}

\caption{\tablabel{overlap}Cosine similarity of ngrams that
  cross word boundaries and 
  disambiguate polysemous words.
  The tables show three disambiguating ngrams
for ``exchange'' and ``rates'' that have different meanings
as indicated by low cosine similarity. In phrases like ``floating
exchange rates'' and ``historic exchange rates'',
disambiguating ngrams overlap.
Parts of the word ``exchange'' are disambiguated by
preceding context (ic@exchang, ing@exchan) and parts of
``exchange'' provide context for disambiguating
``rates'' (xchange@ra).}
\end{table}

\section{Discussion}
\seclabel{discuss} \textbf{Single vs.\ multiple
  segmentation.} The motivation for multiple segmentation is
\emph{exhaustive coverage of the space of possible
segmentations}. An alternative approach would be to attempt
to find a \emph{single optimal segmentation}.

Our intuition is that in many cases \emph{overlapping
  segments contain complementary information}.
\tabref{overlap} gives an example.
Historic exchange rates are different from floating
exchange rates and this is captured by the
low similarity of the ngrams \texttt{ic@exchang} and \texttt{ing@exchan}.
Also, the meaning of ``historic'' and ``floating''
is noncompositional: these two words take on a specialized
meaning in the context of exchange rates. The same is true
for ``rates'': its meaning is not its general meaning in the
compound ``exchange rates''. 
Thus, we need a representation
that contains overlapping segments, so that ``historic'' /
``floating'' and ``exchange'' can disambiguate each other
in the first part of the  compound
and ``exchange'' and ``rates'' can disambiguate
each other in the second part of the compound. A single
segmentation cannot capture these overlapping ngrams.

\textbf{What text-type  are
  tokenization-free approaches most promising for?}
The reviewers thought that language
and text-type
were badly chosen for this paper. Indeed, a morphologically complex
language like Turkish and a noisy text-type like Twitter
would seem to be better choices for a paper on robust text
representation.

However, robust word representation methods like FTX
are effective for 
\emph{within-token} generalization, in particular, effective for
both complex morphology and 
OOVs. If linguistic variability and noise only occur on the
token level, then a tokenization-free approach has fewer
advantages. 

On the other hand, the foregoing discussion of \emph{cross-token}
regularities and disambiguation applies to well-edited
English text as much as it does to other languages and other
text-types as the example of ``exchange'' shows (which is
disambiguated by prior context and provides disambiguating
context to following words) and as is also exemplified by
lines 5--9 in \tabref{figment} (right).

Still, this paper does not directly evaluate the different
contributions that within-token character ngram embeddings
vs.\ cross-token character ngram embeddings make, so this is
an open question. One difficulty is that few corpora are
available that allow the separate evaluation of whitespace
tokenization errors; e.g., OCR corpora generally do not
distinguish a separate class of
whitespace tokenization errors.

\textbf{Position embeddings vs.\ phrase/sentence embeddings.}
Position embeddings may seem to stand in opposition
to phrase/sentence embeddings. For many tasks, we
need a fixed length 
representation of a longer sequence; e.g.,
sentiment analysis models
compute a
fixed-length representation to classify a sentence 
as positive / negative.

To see that position embeddings are compatible
with fixed-length embeddings,
observe first that, in principle, there is \emph{no difference
between word embeddings and position embeddings}
in this respect. Take a sequence that consists of, say, 6
words and 29 characters.
The
initial representation of the sentence has length 6
for word embeddings and length
29 for position embeddings. In both cases, we need a
model that reduces the variable length sequence into a fixed
length vector at some intermediate stage and then classifies
this vector as positive or negative. For example, both
word and position embeddings can be used as the input to an
LSTM whose final hidden unit activations are
a fixed length vector of this type.

So  assessing position embeddings is not a question of
variable-length vs.\ fixed-length representations. Word
embeddings give rise to variable-length representations
too. The question is solely whether the position-embedding
representation is a more effective representation. 

A more specific form of this argument concerns architectures
that compute fixed-length representations of subsequences on
intermediate levels, e.g., CNNs. The difference between
position-embedding-based CNNs and  word-embedding-based CNNs
is that the former have access to a \emph{vastly increased range of
subsequences}, including substrings of words (making it
easier to learn that ``exchange'' and ``exchanges'' are related)
and cross-token character strings  (making it easier to learn
that ``exchange rate'' is
noncompositional). Here, the questions are: (i) how useful are
subsequences made available by position
embeddings  and (ii) is the
increased level of noise 
and decreased efficiency caused by
many useless subsequences
worth the information gained by
adding  useful subsequences.

\textbf{Independence of training and utilization.}
We note that our proposed training and
utilization methods are
completely independent. Position embeddings can be computed
from any set of character-ngram-embeddings (including FTX)
and our character ngram learning algorithm could be used for
applications other than position embeddings, e.g., for
computing word embeddings.

\textbf{Context-free vs.\ context-sensitive embeddings.}
Word embeddings are context-free: a given word $w$ like ``king''
is represented by the same embedding independent of the
context in which $w$ occurs. Position embeddings are context-free as well: if
the maximum size of a character ngram is $k\dnrm{max}$, then the
position embedding of the center of a string $s$ of length
$2k\dnrm{max}-1$ is the same independent of the context in
which $s$ occurs.

It is conceivable that text representations could be
context-sensitive.  For example, the hidden states of a
character language model have been used as a kind of
nonsymbolic text representation
\cite{chrupala13segmentation2,evang13elephant2,chrupala14tweets2}
and these states are context-sensitive.  However, such
models will in general be a second level of representation; e.g., the hidden states of a character
language model generally use character embeddings as
the first level of representation. Conversely, position
embeddings can also be the basis for a context-sensitive
second-level text representation. We have to start somewhere
when we represent text. Position embeddings are motivated by
the desire to provide a representation that can be computed
easily and quickly (i.e., without taking context into
account), but that on the other hand is much richer than the
symbolic alphabet.

\textbf{Processing text vs.\ speech vs.\ images.} \newcite{gillick16} write:
``It is worth noting that noise is often added \ldots\
to images \ldots\ and speech 
where the added noise does not fundamentally
alter the input, but rather blurs it. [bytes allow us to achieve]
something like blurring with text.'' 
It is not clear to what extent blurring on the byte
level is useful; e.g., if we blur the
bytes of the word ``university'' 
individually, then it is unlikely that the noise generated
is helpful in, say, providing good training
examples in parts of the space that would otherwise be unexplored.
In contrast, the
text representation we have introduced in this paper can be
blurred in a way that is analogous to images and
speech. Each embedding of a position is a vector that can be
smoothly changed in every direction. We have showed that the
similarity in this space gives rise to natural variation.

\textbf{Prospects for completely tokenization-free processing.}
We have focused on whitespace tokenization  and
proposed a whitespace-tokenization-free method that computes
embeddings of higher quality than tokenization-based
methods. However, there are many properties of edited text 
beyond whitespace
tokenization
that a complex
rule-based tokenizer exploits. In a small explorative experiment, we replaced
all non-alphanumeric characters with whitespace and repeated experiment A-ORIGINAL for this
setting. This results in an $F_1$ of .593, better by .01 
than the best tokenization-free method. 
This illustrates that there is still a lot of work to be
done before we can obviate
the need for tokenization.

\onlyincludeinlongversion{\section{Related
    work\footnote{This section was written in September 2016
      and revised in April 2017. To suggest
corrections and additional references, please send mail
to inquiries@cislmu.org}}

In the following, we will present an overview of work on
character-based models for a variety of tasks from different NLP areas.\footnote{In our view, 
morpheme-based models are not true instances
  of character-level models as linguistically motivated morphological segmentation is
  an equivalent step to tokenization, but on a different
  level. We therefore do not cover most work on
  morphological segmentation in this paper.}

The history of character-based research in NLP is long and spans a broad array of  tasks. Here we make an attempt to categorize the literature of character-level work  into three classes based on the way they incorporate character-level information into their computational models. The three classes we identified are: \textbf{tokenization-based models}, \textbf{bag-of-n-gram models} and \textbf{end-to-end models}. However, there are also
mixtures possible, such as tokenization-based bag-of-n-gram models
or bag-of-n-gram models trained end-to-end.

On top of the categorization based on the underlying representation model, we sub-categorize the work within each group into six abstract types of NLP tasks (if possible) to be able to compare them more directly. These task types are the following:

\begin{enumerate}

\item \textbf{Representation learning for character sequences:}
Work in this category attempts to learn a generic representation for sequences of characters in
an unsupervised fashion on large corpora. Learning such representations has been shown to be useful for 
solving downstream NLP tasks \cite{collobert11scratch,kocmi16fasttext,mikolov13word2vec}. 

\item \textbf{Sequence-to-sequence generation:} This
  category includes a variety of NLP tasks mapping
  variable-length input sequences to variable-length output
  sequences. 
Tasks in this
  category include those that are naturally suited for character-based
  modeling, such as grapheme-to-phoneme conversion
  \cite{bisani08joint,kaplan1994regular,sejnowski1987parallel},
  transliteration
  \cite{li2004joint,knight1998machine,sajjad2012statistical}, 
spelling normalization for historical text \cite{pettersson2014nmtfornormalization},
  or diacritics restauration \cite{mihalcea02letterlevel}.
Machine translation and question answering
  are other major examples of this category. 

\item \textbf{Sequence labeling:} NLP tasks that assign a categorical label to a part of a sequence  (a character, a sequence of characters or a token) are included within this group. Part-of-speech tagging, named entity recognition, morphological segmentation and word alignment are exemplary instances of sequence labeling.

\item \textbf{Language modeling:} The other type of task for that character-based modeling has been important
for a very long time is language modeling. In 1951, Shannon  \cite{shannon1951prediction} proposed a guessing game asking ``How well can the
next letter of a text be predicted when the preceding N letters are known?''
This is basically the task of character-based language modeling.

\item \textbf{Information retrieval:}
The information retrieval task is to retrieve the most
relevant character sequence to a given character sequence
(the query) from a set of existing  character sequences.

\item \textbf{Sequence classification:}
In this type of NLP task, a categorical label will be assigned to a character sequence (e.g., a document). Instances of this type are language identification, sentiment classification, authorship attribution, topic classification and word sense disambiguation.

\end{enumerate}

\subsection{Tokenization-based Approaches}
We group character-level models that are based on tokenization  as a necessary preprocessing step in the category of tokenization-based approaches. Those can be either models with tokenized text as input
or models that operate only on individual tokens (such as studies on 
morphological inflection of words). 

In the following paragraphs, we cover a subset of
tokenization-based models that are used for representation learning, sequence-to-sequence generation, sequence labeling, language modeling, and sequence classification tasks.

\newparagraph{Representation learning for character sequences} 
Creating word representations based on characters has attracted much attention
recently. Such representations can model rare words, complex words, out-of-vocabulary words and noisy texts. In comparison to traditional word representation models that learn separate vectors for word types, character-level models are more compact as they only need vector representations for characters as well as a compositional model. 

Various neural network architectures have been proposed for
learning token representations based on characters.
Examples of such architectures are 
averaging character embeddings, (bidirectional) recurrent
neural networks (RNNs) (with or without gates) over
character embeddings and convolutional neural networks
(CNNs) over character embeddings. Studies on the general
task of learning word representations from characters
include
\cite{chen15jointcharword,ling15characternmt,luong13better,vylomova16wordrepresentation}. 
These character-based word representations are often combined
with word embeddings and integrated into a hierarchical
system, such as hierarchical RNNs or CNNs
 or combinations of both
to solve other
task types. We will provide more concrete examples in the following paragraphs.  

 \newparagraph{Sequence-to-sequence generation (machine translation)}
Character-based machine translation is no new topic. 
Using character-based methods has been a natural way to overcome 
challenges like rare words or out-of-vocabulary words in machine translation. 
Traditional machine translation models based on characters or character n-grams have been 
investigated by \cite{lepage05purest,tiedemann13characterlevel,vilar07letters}. 
Neural machine translation with character-level and subword units has become popular 
recently \cite{costajussa16characternmt,luong16openvoc,sennrich16bep,vylomova16wordrepresentation}. In such neural models, using a joint attention/translation model makes joint learning of alignment and translation possible \cite{ling15characternmt}.

Both hierarchical RNNs \cite{ling15characternmt,luong16openvoc} 
and combinations of 
CNNs and RNNs  have been proposed for neural machine 
translation \cite{costajussa16characternmt,vylomova16wordrepresentation}.

 \newparagraph{Sequence labeling} 
Examples of early
efforts on sequence labeling using tokenization-based
models
include:
bilingual character-level
alignment extraction \cite{church93charalign}; unsupervised
multilingual part-of-speech induction based on characters
\cite{clark03combining}; part-of-speech tagging with subword/character-level
information \cite{andor16globally,hardmeier16neuralhistorical,ratnaparkhi1996maximum}; 
morphological segmentation and tagging
\cite{cotterell16canonical,muller13higherorder}; and
identification of language inclusion with character-based
features \cite{alex05inclusions}.

Recently, various hierarchical character-level
neural networks have been applied to a variety of sequence
labeling tasks.
\begin{itemize}
\item Recurrent neural networks are used for part-of-speech tagging
\cite{ling2015finding,plank16lstmpostagging,yang2016multitask},
named entity recognition
\cite{lample16neuralner,yang2016multitask},
chunking \cite{yang2016multitask} and morphological
segmentation/inflection generation
\cite{cao16joint,faruqui16seqtoseq,kann16neural,kann16sigmorphon,kann16med,rastogi16neuralcontext,wang16morphological,yubuysblunsom16segmenttosegment}. Such
hierarchical RNNs are also used for dependency parsing
\cite{ballesteros15lstms}. 
This work has shown that morphologically
rich languages  benefit from character-level models in dependency parsing. 

\item Convolutional neural networks are used for part-of-speech
tagging 
\cite{dossantos14postagging} and named entity recognition
\cite{dossantos15boosting}. 

\item The combination of RNNs and CNNs
is used, for instance, for named entity recognition.
\end{itemize}

 \newparagraph{Language modeling}
Earlier work on sub-word language modeling has used morpheme-level features for language models 
\cite{bilmes03factored,ircing2001large,kirchhoff2006morphology,shaik13sublexical,vergyri2004morphology}. In addition, 
hybrid word/n-gram language models for out-of-voca\-bu\-lary words have been applied to speech recognition
\cite{hirsimaki2006unlimited,kombrink2010recovery,parada2011learning,shaik2011hybrid}. Furthermore, characters and
character n-grams have been used as input to restricted boltzmann machine-based language models
for machine translation \cite{sperr13letterngram}.

More recently, character-level neural language modeling has been
proposed by a large body of work
\cite{bojanowski15rnn,botha2014compositional,kim16character,ling2015finding,mikolov12subword,shaik13sublexical,sperr13letterngram}. Although
most of this work is using RNNs, there exist
architectures that combine CNNs and RNNs
\cite{kim16character}. While most of these studies combine
the output of the character model with word embeddings,  the
 authors of \cite{kim16character} report that this does not help
them for their character-aware neural language model. They use
convolution over character embeddings followed by a highway network \cite{highwayNetworks15}
and feed its output into a long short-term memory network
that predicts the next word using a softmax function.

 \newparagraph{Sequence classification}
Examples of
tokenization-based 
models that perform sequence classification are
CNNs used for sentiment classification
\cite{dossantos14cnnsentiment} and combinations of
RNNs and CNNs used for language identification
\cite{jaech16langid}.

\subsection{Bag-of-n-gram Models}
Character n-grams have a long history as features
for specific NLP applications, such as information retrieval.
However, there is also work on representing words or larger input units,
such as phrases, with character n-gram embeddings. Those embeddings
can be within-token or cross-token, i.e., there is no tokenization
necessary.

Although such models learn/use character n-gram embeddings from tokenized text or short text segments, to represent a piece of text, the occurring 
character n-grams are usually summed without the need for tokenization. For example,  the phrase ``Berlin is located in Germany'' is represented with character 4-grams as follows: ``Berl erli rlin lin\_ in\_i n\_is \_is\_ is\_l
s\_lo \_loc loca ocat cate ated ted\_ ed\_i d\_in \_in\_ in\_G n\_Ge \_Ger
Germ erma rman many any.'' Note that the input has not been tokenized and there are n-grams spanning token boundaries. We also include non-embedding approaches using bag-of-n-grams within this group as they go beyond word and token representations.

In the following, we explore a subset of
bag-of-ngram models that are used for representation learning, information retrieval, and sequence classification tasks.

 \newparagraph{Representation learning for character sequences}
An early study in this category of character-based models is  \cite{schutze92wordspace}.
Its goal is to create corpus-based fixed-length distributed semantic representations for text.
To train k-gram embeddings, the top character k-grams are extracted from a corpus
along with their cooccurrence counts. Then, singular value decomposition (SVD) is used to create low dimensional
k-gram embeddings given their cooccurrence matrix.
To apply them to a piece of text, the k-grams of the text are extracted and
their corresponding embeddings are summed. The study evaluates the k-gram embeddings
in the context of word sense disambiguation.

A more recent study  \cite{wieting16charagram} trains character n-gram embeddings in an end-to-end fashion with a neural network. They are evaluated on word similarity,
sentence similarity and part-of-speech tagging.

Training character n-gram embeddings has also been proposed for
biological sequences \cite{asgari15biological,asgari16fifty} for a variety of bioinformatics tasks.

 \newparagraph{Information retrieval}
As mentioned before, character n-gram features are
widely used in the area of information retrieval 
\cite{cavnar95using,chen1997chinese,damashek95ngrams,de1974experiments,mcnamee04character,kettunen10syllables}.

 \newparagraph{Sequence classification}
Bag-of-n-gram models are used for language identification 
 \cite{baldwin10languageid,dunning94languageid}, topic labeling  \cite{kou15trigram}, 
 authorship attribution  \cite{peng2003language}, word/text similarity 
 \cite{bojanowski17enriching,eyecioglu16asobek,wieting16charagram} and word sense disambiguation 
 \cite{schutze92wordspace}.

\subsection{End-to-end Models}
Similar to bag-of-n-gram models, end-to-end models are tokenization-free. Their
input is a sequence of characters or bytes and they are directly optimized on a
(task-specific) objective. Thus, they learn their own, task-specific representation
of the input sequences. Recently, character-based end-to-end models have gained a lot of popularity due to the success of
neural networks. 

We explore the subset of these models that
are used for sequence generation, sequence labeling,
language modeling and sequence classification tasks.

\newparagraph{Sequence-to-sequence generation}
In 2011, the authors of  \cite{sutskever11generating} already proposed an end-to-end model
for generating text. They train RNNs with multiplicative connections on the task
of character-level language modeling. Afterwards, they use the model to generate text and
find that the model captures linguistic structure and a large vocabulary. It produces
only a few uncapitalized non-words and is able to balance parantheses and quotes even
over long distances (e.g., 30 characters).
A similar study by \cite{graves13genrnn} uses a long short-term memory network to
create character sequences.

Recently, character-based neural network
sequence-to-sequence models have been applied to instances of
generation tasks like machine translation
\cite{chung16characternmt,kalchbrenner16lineartime,lee16charlevel,wu16gaphuman,yang-EtAl:2016:COLING} (which was
previously proposed on the token-level
\cite{sutskever2014sequence}), question answering
\cite{golub16character} and speech recognition \cite{bahdanau16endtoend,chan16listenspell,eyben09speechtoletters,graves14endtoend}. 
%and summarization \cite{yubuysblunsom16segmenttosegment}.

 \newparagraph{Sequence labeling} 
Character and character n-gram-based features were already
proposed in 2003 for named entity recognition in an
end-to-end manner using a hidden markov model
\cite{klein03characterlevel}. More recently, the authors of \cite{mahovy16end2end} have proposed 
an end-to-end neural network based model for named entity recognition and part-of-speech tagging. 
An end-to-end model is also suggested for unsupervised, language-independent identification of phrases or 
words  \cite{gerdjikov2016corpus}.

A prominent recent example of neural end-to-end sequence labeling is the paper 
by  \cite{gillick16} about multilingual language processing from bytes. A window is slid over the input sequence, which is represented by its byte string. Thus, the segments in the window can
begin and end mid-word or even mid-character. The authors apply the same
model for different languages and evaluate it on part-of-speech tagging and
named entity recognition.

 \newparagraph{Language modeling}
The authors of \cite{chung16multiscale} propose a hierarchical multiscale recurrent neural
network for language modeling. The model uses different timescales to encode temporal
dependencies and is able to discover hierarchical structures in a character sequence without
explicit tokenization. Other studies on end-to-end language models 
include \cite{kalchbrenner16lineartime,miyamoto16wordcharacter}.

 \newparagraph{Sequence classification}
Another recent end-to-end model uses character-level inputs for document 
classification  \cite{xiao16characterlevel,zhang15scratch,zhang15characterlevel}. To capture long-term dependencies of the input, the authors combine
convolutional layers with recurrent layers. The model is evaluated on sentiment
analysis, ontology classification, question type classification and news categorization.

End-to-end models are also used for entity typing based on the character
sequence of the entity's name  \cite{yaghoobzadeh17multilevel}.

}

\section{Conclusion}
We introduced the first generic text representation model that is
completely nonsymbolic, i.e., it does not require the
availability of a segmentation or tokenization method that
identifies words or other symbolic units in text.
This is true for the  training of the model 
as well
as for applying it when computing the representation of a new text.
In contrast to prior work that has assumed that the sequence-of-character
information captured by character ngrams is sufficient,
position embeddings also capture sequence-of-ngram information.
We showed that our model performs better than prior work
on entity typing and text denoising.

\onlyincludeinlongversion{
\textbf{Future work.}

The most important challenge that we need to address is how
to use nonsymbolic text representation for tasks that are
word-based like part-of-speech tagging. This may seem like a
contradiction at first, but \newcite{gillick16} have
shown how character-based methods can be used for
``symbolic'' tasks. We are currently working on creating an
analogous evaluation for our nonsymbolic text representation.
}

\onlyincludeinshortversion{

\bibliography{buecher}
\bibliographystyle{eacl2017}

}

\onlyincludeinshortversion{

\newpage

\

\newpage

\appendix

\section{Supplementary material}

\subsection{Related work}
The related work section appears in the long version of this paper
\cite{schuetze16nonsymboliclong2}.
}

\subsection{Acknowledgments}
This work was supported by
DFG (SCHUE 2246/10-1) and Volkswagenstiftung. 
We are grateful for their comments to:
the anonymous reviewers, Ehsan Asgari, Annemarie
Friedrich, Helmut Schmid, Martin Schmitt and Yadollah Yaghoobzadeh.

\onlyincludeinlongversion{\section{Appendix}}

\begin{figure}
\includegraphics[width=0.3\textwidth]{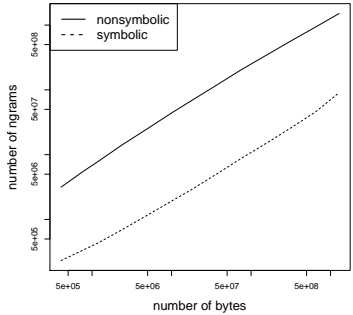}
\caption{The graph shows how many different character ngrams
  ($k\dnrm{min}=3$, 
$k\dnrm{max}=10$) occur in the first $n$ bytes of the
  English Wikipedia for symbolic (tokenization-based)
  vs.\ nonsymbolic (tokenization-free) processing. 
The
  number of ngrams is an order of magnitude larger in the
  nonsymbolic approach. We counted all segments,
  corresponding to $m=\infty$. For the experiments in the
  paper ($m=50$), the number of nonsymbolic character ngrams
  is smaller.
\figlabel{sparseness}}
\end{figure}

\subsection{Sparseness in tokenization-free approaches}
Nonsymbolic representation learning
does not preprocess the training corpus by means of
tokenization and considers many ngrams that would be ignored
in tokenized approaches because they span token
boundaries. As a result, the number of ngrams that occur in a corpus is an
order of magnitude larger for tokenization-free approaches
than for tokenization-based approaches. 
See \figref{sparseness}.

\subsection{Experimental settings}
\textbf{W2V hyperparameter settings.}
		size of word vectors: 200,
		max skip length between words: 5,
		threshold for occurrence of words:
0,
		hierarchical softmax: 0,
		number of negative examples: 5,
		threads: 50,
		training iterations: 1,
	min-count: 5, starting learning rate: .025,
	classes: 0

        \textbf{FTX hyperparameter settings.}
learning rate: .05,
lrUpdateRate: 100, 
size of word vectors: 200,
size of context window: 5,
number of epochs: 1,
minimal number of word occurrences: 5,
number of negatives sampled: 5,
max length of word ngram: 1,
loss function: ns,
number of buckets: 2,000,000,
min length of char ngram: 3,
max length of char ngram: 6,
number of threads: 50,
sampling threshold: .0001

We ran some experiments with more epochs, but this did not
improve the results.

\subsection{Other hyperparameters}
We did not tune  $ N_o=200$, but results are highly
sensitive to the value of this parameter. If $N_o$ is too
small, then beneficial conflations (collapse punctuation
marks, replace all digits with one symbol) are not found. 
If $N_o$ is too large, then precision suffers -- in the
extreme case all characters are collapsed into one.

We also did not tune $m=50$, but we do not consider results
to be very sensitive to the value of $m$ if it is reasonably
large. Of course, if a larger range of character ngram
lengths is chosen, i.e., a larger interval
$[k\dnrm{min},k\dnrm{max}]$, then at some point $m=50$ will
not be sufficient and possible segmentations would not be
covered well enough in sampling.

The type of segmentation used in  multiple segmentation can
also be viewed as a hyperparameter. An alternative to random
segmentation would be exhaustive segementation, but a naive
implementation of that strategy would increase the size of
the training corpus by several orders of magnitude. Another
alternative is to choose one fixed size, e.g., 4 or 5
(similar to \cite{schutze92wordspace}). Many of the nice
disambiguation effects we see in 
\tabref{figment} (right)
and in \tabref{overlap} would not be possible with short
ngrams. On the other hand, a fixed ngram size that is
larger, e.g., 10, would make it difficult to get 100\%
coverage: there would be positions for which no position
embedding can be computed.

\onlyincludeinlongversion{

\bibliography{buecher}
\bibliographystyle{eacl2017}

\appendix

}

\end{document}